\documentclass[11pt]{article}
\usepackage{coling2020}
\usepackage{times}
\usepackage{url}
\usepackage{latexsym}
\usepackage{float}
\usepackage{color}
\usepackage{soul}

\usepackage{marvosym}
\usepackage{ifsym}
\usepackage{graphicx}
\usepackage{microtype}
\usepackage{graphicx}
\usepackage{booktabs} 

\usepackage{adjustbox}
\usepackage{float}

\usepackage{mathtools}
\usepackage{color}
\usepackage{soul}
\usepackage{amsmath}
\usepackage{amssymb}

\usepackage{hyperref}
\usepackage{cleveref}

\colingfinalcopy 

\title{Bayesian Methods for Semi-supervised Text Annotation}

\author{Kristian Miok$^{1,2}$, Gregor Pir\v{s}$^1$ and Marko Robnik-\v{S}ikonja$^1$\\ \\
$^1$ University of Ljubljana, Faculty of Computer and Information Science, Slovenia \\
Email: \{gregor.pirs, marko.robnik\}@fri.uni-lj.si \\
$^2$ West University of Timisoara, Computer Science Department, Romania \\
Email: kristian.miok@e-uvt.ro
}

\begin{document}

\maketitle
\begin{abstract}
Human annotations are an important source of information in the development of natural language understanding approaches. As under the pressure of productivity annotators can assign different labels to a given text, the quality of produced annotations frequently varies. This is especially the case if decisions are difficult, with high cognitive load, requires awareness of broader context, or careful consideration of background knowledge. To alleviate the problem, we propose two semi-supervised methods to guide the annotation process: a Bayesian deep learning model and a Bayesian ensemble method. Using a Bayesian deep learning method, we can discover annotations that cannot be trusted and might require reannotation. A recently proposed Bayesian ensemble method helps us to combine the annotators' labels with predictions of trained models. According to the results obtained from three hate speech detection experiments, the proposed Bayesian methods can improve the annotations and prediction performance of BERT models.
\end{abstract}

\section{Introduction}
\label{intro}

Recent successful applications of artificial intelligence in various fields, including natural language processing, are often due to long hours of human annotation when preparing datasets for machine learning. The annotation process transfers human knowledge to machine learning models but it is often done under time pressure and with inadequate instructions or with insufficiently trained annotators.Aiming to make the annotation process easier, we study the possibility of designing a data labeling process which requires less human supervision.

In practice, a fairly standard procedure in the annotation quality control is to recheck the labels that are wrongly classified by using several prediction models. As an alternative, Bayesian inference produces a distribution of possible decisions and can improve the selection  of  instances requiring reannotation \cite{miok2020ban}. Most neural networks do not support the assessment of predictive uncertainty. The Bayesian inference framework can be helpful, however, most techniques do not scale well in neural networks with high dimensional parameter space \cite{izmailov2019subspace}. Various methods were proposed to overcome this problem \cite{myshkov2016posterior}, one of the most efficient being Monte Carlo Dropout (MCD) \cite{gal2016dropout}. Its idea is to use the dropout mechanism in neural networks as a regularization technique \cite{srivastava2014dropout} and interpret it as a Bayesian optimization approach that samples from the approximate posterior distribution.

A common problem in text annotations is that annotators are not always sure about correct labels  due to  uncertainty in the text \cite{vincze2015uncertainty,szarvas2008bioscope}. On difficult texts, annotators frequently give ambiguous labels and their annotations can be biased. 
Instead of asking annotators to label the raw text, it would be easier for them if they were proposed answers accompanied by probabilistic scores from an ensemble of predictive models. Ensemble methods produce robust models that frequently provide significantly better predictions than individual models.  The key strength of ensembles is that they can overcome errors and shortcomings of individual ensemble members. However,  diversity in combining different predictions and reliability of individual predictions need to be better understood and evaluated \cite{zhou2012ensemble}. A recently published ensemble method Multivariate Normal Mixture Conditional Likelihood Model (MM) \cite{pirs2019bayesian} tries to understand the predictors on the distributional level and use Bayesian inference to combine them. In this work, we evaluate MM's performance when combining predictive models on the hate speech detection task. We show that our methodology can serve as a helpful tool in the data annotation process.  

Recently, the most successful approach in text classification is to use transformer neural networks  \cite{vaswani2017attention}, pretrained on large monolingual corpora, and then fine-tune them for a specific task, such as text classification. For example, BERT (Bidirectional Encoder Representations from Transformers) \cite{devlin2019-bert} uses masked language modeling and order of  sentences prediction tasks  to  build a general language understanding model. During the fine-tuning for a specific downstream task, additional layers are added to the BERT model, and the model is trained on the data of interest to capture the specific knowledge required to perform the task. 


The main aims of the paper is to propose methods that can save time and resources during the text annotation process and improve prediction performance.
As a test domain we use hate speech detection in tweets, news comments and Facebook comments. We investigate two performance improving techniques which can be summarized as our main contributions as follows.
\begin{enumerate}
    \item We remove instances with uncertain classifications from the training set and show that fine-tuning on the cleaned dataset improves the performance of the BERT model. Less certain classifications can be selected for reannotation.
    \item We combine predictions of machine learning models using the MM probabilistic ensemble method. The approach is beneficial for predictive performance.   
\end{enumerate}


The paper consists of five further sections. In Section 2, we present related works on prediction uncertainty and hate speech detection. In Section 3, we propose the methodology for uncertainty assessment of deep neural networks using attention layers and MCD. In Section 4, we describe the tested datasets and evaluation scenarios. The obtained results are presented in Section 5, followed by conclusions and ideas for further work in Section 6.

\section{Related Work}

In this section, we introduce related work split into four topics. First, we present the work on semi-supervised learning that can be used in text annotation, followed by the related research on Bayesian learning for text classification. In the third subsection, we describe probabilistic ensemble methods and in the fourth, we outline the related  work on hate speech detection.

\subsection{Semi-supervised Learning for Text Annotation}
The performance of supervised learning depends on the availability of a sufficient amount of labeled data. However, manual labeling is expensive and difficult to scale up to large amounts of data. Semi-supervised learning tries to utilize large amounts of unlabeled data available for many problems by combining them with small amounts of labeled data \cite{zhu2005semi}. The goal of semi-supervised learning is to understand how combining labeled and unlabeled data can change the learning behavior, and design algorithms that take advantage of such a combination \cite{zhu2009introduction}. Most semi-supervised learning strategies extend either unsupervised or supervised learning to include additional information typical of the other learning paradigm. The transductive learning is related to the semi-supervised learning, but assumes that the test set is known in advance and its goal is to optimize the generalization ability on this (unlabeled) test set \cite{zhou2010semi}. In the non-transductive setting, \newcite{acharya2013probabilistic} combine probabilistic classifiers. They take class labels from existing classifiers and cluster labels from a clustering ensemble. The consensus labeling is assigned to the target data.

\subsection{Bayesian Methods for Text Classification}

Although, recent works on prediction uncertainty mostly investigate deep neural networks, many other probabilistic classifiers were analyzed in the past \cite{Platt99probabilisticoutputs,Niculescu05,zhang2013robust,cao2015classification,he2018knowledge}. Prediction reliability is an important issue for black-box models like neural networks as they do not provide interpretability or reliability information about their predictions. Most existing reliability scores for deep neural networks are contructed using Bayesian inference. The most popular exception is the work of \newcite{lakshminarayanan2017simple}, who proposed to use deep ensembles to estimate the prediction uncertainty. 

A computationally efficient simulation of Bayesian inference uses Monte Carlo dropout \cite{gal2016dropout}. The first implementation of dropout in recurrent neural networks (RNNs) was in 2013 \cite{wang2013fast} but further research revealed a negative impact of dropout in RNNs \cite{bluche2015apply}. Later, the dropout was successfully applied to language modeling by \newcite{zaremba2014recurrent} who used it only in fully connected layers. \newcite{gal2016theoretically} implemented the variational inference based dropout which can regularize also recurrent layers. In this way method mimics Bayesian inference by combining probabilistic parameter interpretation and deep RNNs. Several other works investigate how to estimate prediction uncertainty within different data frameworks using RNNs \cite{zhu2017deep,miok2019prediction}, e.g., Bayes by Backpropagation (BBB) was applied to RNNs \cite{fortunato2017bayesian}. Monte Carlo dropout was also introduced into variational autoencoders \cite{miok2019,miok2019generating} and for estimating prediction intervals \cite{miok2018estimation}.

To our knowledge, Bayesian deep learning models were not yet used to detect less certain text classifications and remove them from a train dataset to improve the prediction performance.

\subsection{Probabilistic Ensembles}
Most methods used for text classification can produce probabilistic predictions which are rarely exploited beyond classification into a discrete class. As probabilistic predictions provide additional information compared to the discrete outcome, we use ensembles that can model predictive distributions. Ensemble methods can be divided into two main groups. The first group of methods estimates the performance of individual classifiers and weights them accordingly. The second group of methods learns the structure of predictions and bases their forecasts on it.

The first group of methods can be further divided into methods that are able to combine full posterior distributions and methods that only combine probabilistic point predictions. The advantage of the former is that they are more expressive, and a disadvantage is that they require inputs in the form of a full distribution, which is not always available. Bayesian model averaging \cite{Hoeting1999} combines models by their marginal posterior probability. This method is suitable if one of the candidate models is the true data generating process, otherwise its performance decreases \cite{Cerquides2005}. Bayesian stacking \cite{Yao2018} is also useful in these cases. It is based on weighing the posterior predictive distributions of individual models by estimating their leave-one-out cross-validation performance. Linear opinion pool \cite{Cooke1991} is a classical approach to combine classifiers and can be included into the second group of methods. It combines predictions as a linear combination of individual models by maximizing the likelihood. An example of Bayesian approach from the second group is the agnostic Bayesian learning of ensembles \cite{Lacoste2014}, which weighs the models by estimated probabilities of them being the best model; the estimates are based on holdout computation of generalization performance.

Methods that model the structure of predictions are  especially useful in case of complex relationships between individual models' predictions and the response variable. Independent Bayesian classifier combination (IBCC) \cite{Kim2012} combines non-probabilistic predictions by estimating the probability mass of predictions with a categorical distribution, conditional on the true label. It provides probabilities for a new observation that are proportional to the probability mass of new inputs for each true label. \newcite{Nazabal2016} has extended the IBCC to probabilistic predictions by using the Dirichlet distribution. Supra-Bayesian methods \cite{Lindley1985} combine probabilistic predictions using the log-odds of probabilities and modeling them with the multivariate normal (MVN) distribution, conditional on the true label. They use the common covariance matrix over all true labels but vary the means. 

Ensemble modeling has recently been studied also within the text classification area \cite{li2018text,silva2010distributed,kilimci2018deep}, but not within the context of probabilistic ensemble models. In our work, we investigate Bayesian ensemble modeling for hate speech detection and how this can improve individual model predictions.

 \subsection{Hate Speech Detection}

Analyzing sentiments and extracting emotions from the text are  useful natural language processing applications \cite{sun2018emotional}. Being one of the wide range of applications where machines tend to understand human sentiments, hate speech detection is gaining importance with the rise of social media. We regard hate speech as written or oral communication that abuses or threatens a specific group or target \cite{warner2012detecting}. 

Detecting abusive language for less-resourced languages is difficult, hence, multilingual and cross-lingual methods are employed to improve the results \cite{stappen2020cross}. This is especially the case when the involved languages are morphologically or geographically similar \cite{pamungkas2019cross}. In our work, we investigate and compare hate speech detection methods for English, Croatian, and Slovene.  English is by far the most researched language with plenty of resources  \cite{malmasi2017detecting,davidson2017automated,waseem2016hateful}. Recently, hate speech detection studies were done also on neighbouring Slavic languages   Croatian \cite{kocijan2019detecting,ljubevsic2018datasets} and Slovene \cite{fivser2017legal,ljubevsic2019frenk,vezjak2018radical}.  

Hate speech detection is usually treated as a binary text classification problem, and is approached with supervised learning methods. In the past, the most frequently used classifier was the Support Vector Machine (SVM) method \cite{schmidt2017survey}, but recently deep neural networks showed superior performance, first through recurrent neural networks \cite{mehdad2016characters}, and recently using large  pretrained transformer networks  \cite{mozafari2019bert,wiedemann2020uhh}. In this work, we use the recent state-of-the-art pretrained (multilingual) BERT model.

\section{Methods}
We describe two approaches to the assessment of prediction reliability, Bayesian Attention Networks and Bayesian Probabilistic Ensembles.

\subsection{Bayesian Attention Networks}
The work \cite{miok2020ban} that introduce method named `Bayesian Attention Networks' (BAN), proposes the dropout layers to be active also during the prediction phase. In this way, predictions are rather random and are sampled from the \emph{learned} distribution, thereby forming an ensemble of predictions. The obtained distribution can be, for example, inspected for higher moment properties and it can offer additional information on the certainty of a given prediction. During the prediction phase, all layers of the network except the dropout layers are deactivated. The forward pass on such partially activated architecture is repeated for a fixed number of samples, each time producing a different outcome that can be combined into the final probability, or inspected as a probability distribution.

Monte Carlo dropout was adapted for the BERT model in the same way as for BANs. MCD can provide multiple predictions during the test time without any additional training \cite{gal2016uncertainty}. Training a neural network with dropout spreads the information contained in the neurons across the network. Hence, during the prediction, such a trained neural network will be robust; using the dropout principle, a new prediction is created in each forward pass, and a sufficiently large set of such predictions can be used to estimate prediction reliability. The BERT models are trained with 10\% of dropout in all of the layers by default. Therefore, it allows for multiple predictions with the fine-tuned model. We call this model MCD BERT. A possible limitation of this approach is that during training a single dropout rate of 10\% is used, while other dropout probabilities might be more suitable for reliability estimation. We leave this question for further work as it requires long and costly training of several BERT models. 
 
\subsection{Bayesian Probabilistic Ensemble}
To alleviate the drawbacks of individual classification models, we propose the use of MM \cite{pirs2019bayesian}, a Bayesian ensemble method suitable for combining correlated probabilistic predictions. MM is an extension of IBCC \cite{Kim2012}, which combines non-probabilistic predictions. The method is based on finding the latent structure of combined predictions and provides new probabilities based on its distribution. Let $m$ be the number of classes and $r$ the number of individual models we are combining. The main idea is similar to Supra-Bayesian ensembles \cite{Lindley1985}, as we first transform individual probabilistic predictions with the inverse logistic transformation (log-odds) to move from [0,1] space to the $\mathbb{R}$ space. We merge the transformed predictions of individual models and get a $(m - 1) r$-variate distribution. We model this latent distribution with multivariate normal mixtures, conditional on the true label in a similar fashion as in the case of linear discriminant analysis. Let $\theta$ represent estimated parameters and $\theta_t$ the subset of parameters estimated for observations with true label $t$. Let $T^* \in \{1,2,...,m\}$ be the response random variable for a new observation and $u^* \in \mathbb{R}^{(m-1) r}$ the transformed and merged predictions for this new observation. Probabilistic predictions for unseen data can then be generated by calculating the densities of merged predictions for new data:

\begin{equation*}
    p(T^* = t | u^*, \theta) = \frac{p(u^* | \theta_t)(\gamma_t n_t)}{\sum_{i=1}^r p(u^* | \theta_i) (\gamma_i n_i)},
\end{equation*}

\noindent where $p$ is the MVN mixture probability density, $\gamma_t$ is the frequency prior for class $t$, and $n_t$ is the number of true labels in class $t$ in the training dataset. The method uses a regularization term, which increases the variance in any dimension that is difficult to model or has a detrimental effect on the results, effectively decreasing its effect. For a complete Bayesian specification and the derivation of the Gibbs sampler, we refer the reader to \cite{pirs2019bayesian}. We used the same priors as proposed in this paper.

MM is well-suited for combining biased classifiers, or classifiers with systematic errors. It can serve as a calibration tool for an individual classifier by learning its latent distribution. Since BERT is usually accurate but less well calibrated, the MM method has the potential to alleviate miscalibration, while improving or at least preserving the classification performance.

\section{Experimental Setting}
 We first introduce the three phases of  our experiments, followed by the used datasets and implementation details.
The experimental setting consists of three phases: 
\begin{enumerate}
    \item We categorize classifications to trusted and untrusted based on the uncertainty measure from MCD BERT. In this way, we can detect borderline classification that make a false impression of certainty.
    \item We remove the instances with uncertain classifications from the \emph{training set} to improve the dataset on which the BERT model is fine-tuned. This provides better quality data for training and shall improve the quality of the resulting prediction model.
    \item We use Bayesian ensemble to combine automatic predictions with annotators' decisions to remove low-quality training instances. 
\end{enumerate}

\subsection{Datasets}
To test the proposed methodology in the multilingual context, we trained the presented classification models on three different datasets, summarized in Table \ref{table:inputs}.

\begin{enumerate}
    \item The \textbf{English} dataset\footnote{\url{https://github.com/t-davidson/hate-speech-and-offensive-language}} is extracted from the hate speech and offensive language detection study of \newcite{davidson2017automated}. We used the subset of data consisting of 5,000 tweets. We took 1,430 tweets labeled as hate speech and randomly sampled  3,670 tweets from the collection of the remaining 23,353 tweets. 
    
    \item The \textbf{Croatian} dataset was provided by the Styria media company within the EU Horizon 2020  EMBEDDIA project\footnote{\url{http://embeddia.eu}}. The texts were extracted from  user comments in the news portal Ve\v{c}ernji list\footnote{\url{https://www.vecernji.hr}}. The original dataset consists of 9,646,634 comments from which we selected 8,422 comments of which 50\% are labeled as hate speech by human moderators and the other half was randomly chosen from the non-problematic comments.
    
    \item \textbf{Slovene} dataset is a result of the Slovenian national project FRENK\footnote{\url{http://nl.ijs.si/frenk/}  (Research on Electronic Inappropriate Communication)}. Our dataset comes from two studies on Facebook comments \cite{ljubevsic2019frenk}. The first study deals with LGBT homophobia topics while the second analyzes anti-migrants posts. We used all  2,188 hate speech comments, and randomly sampled 3,812 non-hate speech comments. 
\end{enumerate}

\begin{table}[htbp]
\caption{Characteristics of the used datasets: type and number of instances, as well as the input embeddings for each of the datasets.}
\renewcommand{\arraystretch}{1}
\setlength{\tabcolsep}{5pt}
\label{table:inputs}
\centering
\begin{tabular}{lccccc}
    \textbf{Dataset} & \textbf{type} & \textbf{Size}  &  \textbf{Hate} & \textbf{Non-hate} & \textbf{LSTM embeddings}  \\
    \hline
    \textbf{English} & tweets & 5000 & 1430 & 3670 & sentence \\
    \textbf{Croatian} & news comments & 8422 & 4211 & 4211 & fastText \\
    \textbf{Slovene} & Facebook comments & 6000 & 2188  & 3812 & fastText \\
     \hline
\end{tabular}
\end{table}

\subsection{Implementation}
The two Bayesian methods that were proposed in this paper to improve the annotation process have full implementation within their original papers. All of the particularities of how MCD BERT was implemented in PyTorch library\footnote{\url{https://github.com/KristianMiok/Bayesian-BERT}} are presented in \cite{miok2020ban}. The implementation details of the MM method are clearly explained in \cite{pirs2019bayesian} methods section and in this paper we provide full R code \footnote{\url{https://github.com/gregorp90/MM}}.

\section{Results}

In this section, we present three groups of results: removing uncertain instances from the training set, creating a cleaner training set, and improving annotations using he Bayesian ensembles. 

\subsection{Removing Uncertain Instances}

Using MCD BERT, we obtain multiple predictions for each test set instance, and compute their mean and variance. Using the mean, we determine the classification (hate speech or not), while the variance reports on the certainty of the BERT for this specific instance. Based on the variance, we group classifications into certain and uncertain. Unsurprisingly, removing the uncertain test set instances improves the prediction performance as shown in Table \ref{table:test}, but also leaves a portion of borderline instances unclassified.


\begin{table}[htbp]
\caption{Performance of multilingual BERT model, after removing uncertain instances from the test set of 1000 comments.}
\renewcommand{\arraystretch}{1}
\setlength{\tabcolsep}{5pt}
\label{table:test}
\centering
\begin{tabular}{clcccc}
\textbf{Language} &    \textbf{Metric} & \textbf{Full dataset} & \textbf{200 removed} & \textbf{500 removed} & \textbf{700 removed}  \\
    \hline
    &\textbf{Accuracy} & 0.91  & 0.96 & 0.996  & 0.997  \\
\textbf{EN}&    \textbf{Precision} & 0.90 & 0.95  & 0.992  & 0.994 \\
    &\textbf{Recall} & 0.89 & 0.95 & 1  & 1  \\
    &\textbf{F1} & 0.88 & 0.95  & 0.995  &   0.997\\
     \hline
    &\textbf{Accuracy} & 0.72 & 0.76 & 0.84 & 0.87 \\
\textbf{CRO}    &\textbf{Precision} & 0.68 &  0.71 & 0.80 & 0.85 \\
    &\textbf{Recall} & 0.54 & 0.69 & 0.78 & 0.75  \\
    &\textbf{F1} & 0.61 & 0.70 & 0.79 &  0.83  \\
     \hline
    &\textbf{Accuracy} & 0.71 & 0.76 & 0.83 & 0.87  \\
\textbf{SLO}    &\textbf{Precision} & 0.60 & 0.65& 0.70 & 0.65  \\
    &\textbf{Recall} & 0.56 & 0.64   & 0.66 & 0.54  \\
    &\textbf{F1} & 0.58 & 0.65 & 0.68 & 0.59  \\
     \hline
\end{tabular}
\end{table}

From Table \ref{table:test} we can conclude that the variance of MCD BERT predictions is correlated with the performance of models: the more variance there is in the predictions the less accurate the model. Thus, removing the uncertain classifications can seemingly improve the performance of the test set. A practical benefit of this is that uncertain classification could be passed back to annotators to recheck them.   

\subsection{Creating Cleaner Training Sets}
While the removal of uncertain instances from the test set might just sweep the problematic instances under the carpet, a more practical benefit is to use the uncertainty information to create a better training set.
The test tweets/comments were removed based on how variate are their predictions. Thus, we repeatedly train the MCD BERT model on part of the dataset and use this model to obtain multiple predictions on the other part of the training dataset. In such a way, we collect multiple predictions for all original training tweets or comments and remove observations with the highest prediction variance. As a result of this procedure, 15 and 18 percent of the most uncertain predictions were removed for the English and Slovene dataset respectively. Croatian dataset contains a lot of comments with high variability in their predictions so for this dataset we removed around 35\% of the most uncertain comments. The details of how many instances were removed for each of the three datasets are presented in Table \ref{table:removing}.

\begin{table}[htbp]
\caption{Sizes of the datasets before and after the removing: original number of instances, number of instances removed and final training data size.}
\renewcommand{\arraystretch}{1}
\setlength{\tabcolsep}{5pt}
\label{table:removing}
\centering
\begin{tabular}{lcccc}
    \textbf{Dataset} & \textbf{Training Size} & \textbf{Number of removed}  &  \textbf{Final Size} &  \textbf{Percent removed}  \\
    \hline
    \textbf{English} & 4000 & 719 & 3281 & 18 \% \\
    \textbf{Croatian} & 7422 & 2615 & 4807 & 35\% \\
    \textbf{Slovene} & 5000 & 731 & 4269 & 15 \%  \\
     \hline
\end{tabular}
\end{table}

Using prediction certainty to remove the uncertain instances from the training can improve the fine-tuning of BERT. For  neural network models, during training or fine-tuning  their performance is evaluated on a separate validation set. In Table \ref{table:training}, we can observe how the prediction accuracy on the validation set is improved with number of training epochs. We can see that fine-tuning BERT  on the cleaner dataset improves its performance. We hypothesize that when the uncertainty due to unreliable labels is reduced, the decision boundary is easier to determine. 

\begin{table}[H]
\caption{Performance (measured using $F_1$ score) on the validation sets during training for original and cleaned datasets.}
\renewcommand{\arraystretch}{1}
\setlength{\tabcolsep}{5pt}
\label{table:training}
\centering
\begin{tabular}{lcccccc}
     &\multicolumn{2}{c}{\textbf{English}} & \multicolumn{2}{c}{\textbf{Croatian}} & \multicolumn{2}{c}{\textbf{Slovene}}   \\
         & \textbf{Original} & \textbf{Cleaned} & \textbf{Original} & \textbf{Cleaned}  & \textbf{Original} & \textbf{Cleaned}   \\
    \hline
    \textbf{Epoch1} & 0.92 & 0.98   &  0.68 &  0.77 & 0.70 & 0.64\\
    \textbf{Epoch2} & 0.92 & 0.98   & 0.69 &  0.77 & 0.70 & 0.77\\
    \textbf{Epoch3} & 0.92 & 0.98  &  0.68 & 0.78 & 0.71 & 0.79\\
    \textbf{Epoch4} & 0.92 & 0.98  & 0.70  & 0.79 & 0.72 & 0.81\\
     \hline
\end{tabular}
\end{table}

Results for the model fine-tuned on the cleaned dataset are contained in Table \ref{table:cleaned}. Compared to the results in Table \ref{table:test} (see the ''Full dataset'' column), the prediction results for Croatian and Slovenian datasets are improved while for the English dataset this is not the case. We explain this by the fact that the  English dataset is well-annotated with high-quality predictions. On the other hand, we believe that the Croatian and Slovenian datasets are less clean and contain several questionable annotations. This can be confirmed for the Croatian dataset, which was created within the project we participate in, so we are well-informed about the annotation process.

\begin{table}[H]
\caption{Test set performance ($F_1$ score) of the models trained on the cleaned datasets.}
\renewcommand{\arraystretch}{1}
\setlength{\tabcolsep}{5pt}
\label{table:cleaned}
\centering
\begin{tabular}{lccc}
    \textbf{Metrics} & \textbf{English} & \textbf{Croatian} & \textbf{Slovene}   \\
    \hline
    \textbf{Accuracy} & 0.87 & 0.74  &  0.72  \\
    \textbf{Precision} & 0.88 & 0.73  & 0.62  \\
    \textbf{Recall} & 0.81 & 0.60  &  0.55 \\
    \textbf{F1} & 0.85 & 0.66 & 0.59  \\
     \hline
\end{tabular}
\end{table}



\subsection{Improving Annotations using Bayesian Ensembles}
\begin{figure*}[ht]
\begin{minipage}{0.5 \linewidth}
 \includegraphics[width=\linewidth]{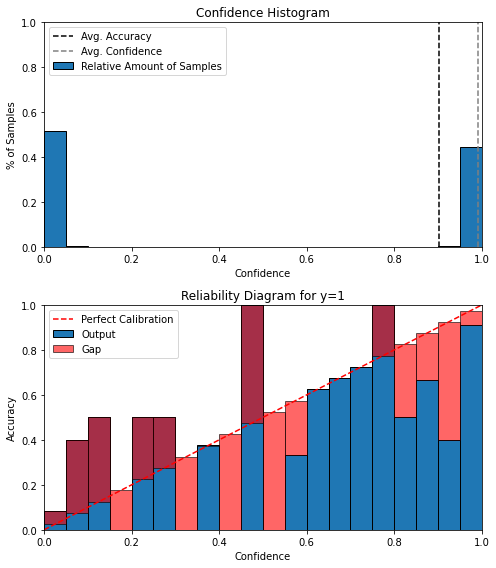}     
\end{minipage}
\hfill
\begin{minipage}{0.5 \linewidth}
 \includegraphics[width=\linewidth]{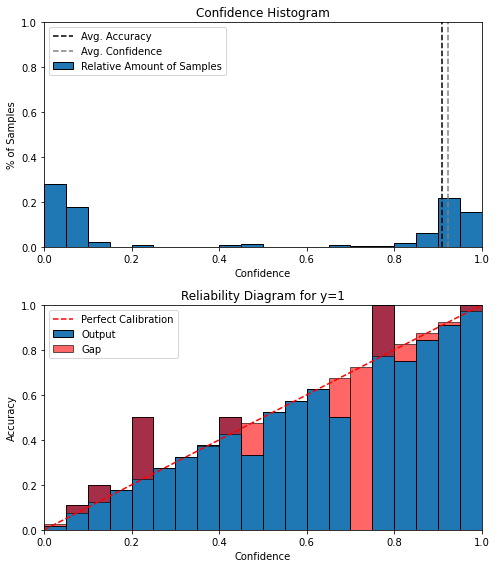}     
\end{minipage}

\caption{Calibration for the BERT predictions (left) and MM model predictions (right).}
\label{fig:Sentic3}
\end{figure*}

We propose a Bayesian ensemble as a support method for the annotation process. As annotators can be distracted, biased, or influenced, we propose to use the MM method to provide them a hint of how shall they annotate the instances. From Table \ref{table:MM}, we can observe that by combining probabilistic predictions of BERT, random forest, and support vector machines, we can further improve the predictive performance. The MM ensemble not only improves BERT's results but also provides better calibrated predictions as evidenced from Figure \ref{fig:Sentic3}.





\begin{table}[H]
\caption{The $F_1$ score of the hate speech classifiers and their ensemble. }
\centering
\label{table:MM}

\begin{tabular}{cccc}

\textbf{Method}  & \textbf{English}& \textbf{Croatian}  & \textbf{Slovene} \\
\hline
\textbf{BERT}		& 0.91 &	0.72	& 0.71  \\
 \textbf{RF}		& 0.83&	0.67		& 0.65 \\
  \textbf{SVM}		& 0.86 &	0.71	& 0.69 \\
 \textbf{MM}  &\textbf{0.92} & \textbf{0.74}    & \textbf{0.72}  \\
\hline
\end{tabular}
\end{table}

\newpage
\section{Conclusions and Further Work}

A large amount of currently available textual data allows and requires modeling with machine learning methods. To apply the supervised methods, the text data has to be annotated, and effective learning requires accurate annotations, which may be expensive for organizers and difficult for human annotators. For this reason, the annotation process is often coupled with semi-supervised machine learning and classification reliability estimation. 

We presented several machine learning approaches, based on Bayesian inference, that can improve the data annotation process. First, multiple predictions obtained with MCD BERT can identify instances with questionable labeling. Second, removing training instances with unreliable labels can improve the quality of the training set, making it more homogeneous and cleaner,  thereby improving the predictive performance of BERT models. Third,  probabilistic ensemble combinations can help annotators to better label the data by providing more accurate and better calibrated prediction probabilities. In conclusion, Bayesian methods can improve the annotation process and shall be further investigated and improved for this task.

In further work, we will focus on improving our method on how to remove uncertain instances. We will construct and test a workflow for semi-supervised text annotation in a real-world setting. Testing different dropout levels in the BERT model may provide a better understanding of its uncertainty and calibration.  

 \section*{Acknowledgements}
This paper was supported by European Union’s Horizon 2020 Programme project EMBEDDIA (Cross-Lingual Embeddings for Less-Represented Languages in European News Media, grant no. 825153). The research was supported  by the Slovenian Research Agency through research core funding no. P6-0411, project 
CANDAS (Computer-assisted multilingual news discourse analysis with contextual embeddings, grant no. J6-2581), and Young researcher grant (Gregor Pir\v s).

\bibliographystyle{coling}
\bibliography{law}

\end{document}